\documentclass[lettersize,journal]{IEEEtran}
\usepackage{amsmath,amsfonts}
\usepackage{algorithmic}
\usepackage{algorithm}
\usepackage{array}
\usepackage[caption=false,font=normalsize,labelfont=sf,textfont=sf]{subfig}
\usepackage{textcomp}
\usepackage{stfloats}
\usepackage{url}
\usepackage{verbatim}
\usepackage{graphicx}
\usepackage{cite}

\usepackage{amssymb}
\usepackage{bbm}

\usepackage[hidelinks]{hyperref}
\hypersetup{hidelinks,
	colorlinks=true,
        linkcolor=black,
        anchorcolor=blue,
        citecolor=blue,
	pdfstartview=Fit,
	breaklinks=true}

\begin{document}

\title{Unsupervised Cross-domain Pulmonary Nodule Detection without Source Data}

\author{Rui Xu,~Yong Luo,~and~Bo Du
\thanks{R. Xu, Y. Luo, and B. Du are with the School of Computer Science, Wuhan University, Wuhan 430072, China (e-mail: rui.xu@whu.edu.cn, yluo180@gmail.com, dubo@whu.edu.cn).}
\thanks{Corresponding authors: Yong Luo and Bo Du.}}



\maketitle

\begin{abstract}
Cross-domain pulmonary nodule detection suffers from performance degradation due to a large shift of data distributions between the source and target domain. Besides, considering the high cost of medical data annotation, it is often assumed that the target images are unlabeled. Existing approaches have made much progress for this unsupervised domain adaptation setting. However, this setting is still rarely plausible in medical applications since the source medical data are often not accessible due to privacy concerns. This motivates us to propose a Source-free Unsupervised cross-domain method for Pulmonary nodule detection (SUP), named Instance-level Contrastive Instruction fine-tuning framework (ICI). It first adapts the source model to the target domain by utilizing instance-level contrastive learning. Then the adapted model is trained in a teacher-student interaction manner, and a weighted entropy loss is incorporated to further improve the accuracy. We establish a benchmark by adapting a pre-trained source model to three popular datasets for pulmonary nodule detection. To the best of our knowledge, this represents the first exploration of source-free unsupervised domain adaptation in medical image object detection. Our extensive evaluations reveal that SUP-ICI substantially surpasses existing state-of-the-art approaches, achieving FROC score improvements ranging from 8.98\% to 16.05\%. This breakthrough not only sets a new precedent for domain adaptation techniques in medical imaging but also significantly advances the field toward overcoming challenges posed by data privacy and availability. Code: \url{https://github.com/Ruixxxx/SFUDA}.
\end{abstract}


\section{Introduction}
\label{sec:intro}

\IEEEPARstart{D}{eep} learning has achieved remarkable success in various object detection tasks. In the medical field, deep networks are able to reach clinical expert-level performance, e.g. pulmonary nodule detection \cite{nodulenet,21PAMI-SANet}, etc. Nonetheless, these networks are usually domain-specific. In other words, they work well when the training/source and target data are drawn from similar distributions. However, in real-world applications, data from different medical centers or scenarios often have different distributions. The networks obtained from the source domain usually have significant performance degradation on the target datasets, which impedes the applications of the deep learning algorithms in real-world medical image analysis. Furthermore, the medical data are time-consuming and expensive to annotate, leading to the lack of labeled training samples for the target domain. Therefore, object detection under the setting of unsupervised domain adaptation (UDA) has gotten a lot of attention in recent years, and considerable effort has been devoted to this problem setting. Since there exists domain shifts between the source and target domain in terms of the illumination, background, style, object appearance, etc, many works apply feature alignment to reduce the domain gap \cite{cvpr18/DA_Faster_RCNN,cvpr19/Strong_Weak_DA,iccv19/Multi_adversarial_Faster_RCNN,iccv19/Multi-level_DA,cvpr20/Cate_regu_DA,eccv20/Center_Aware,cvpr20/coarse2fine,cvpr21/rpn_align}. Some other works regard the UDA task as training with noisy labels on the target domain \cite{iccv19/Robust_DA,cvpr19/Mean_T_DA,iccv19/Self-training_One,eccv20/tri-way_DA,eccv20/Colla_DA,cvpr21/unbiased_T,cvpr22/Target-P_Distill}. 
Existing UDA object detection methods require access to samples from the source domain. This often involves using labeled source images for feature alignment or to ensure training stability.

\begin{figure}[!t]
  \centering
  \includegraphics[width=3.5in]{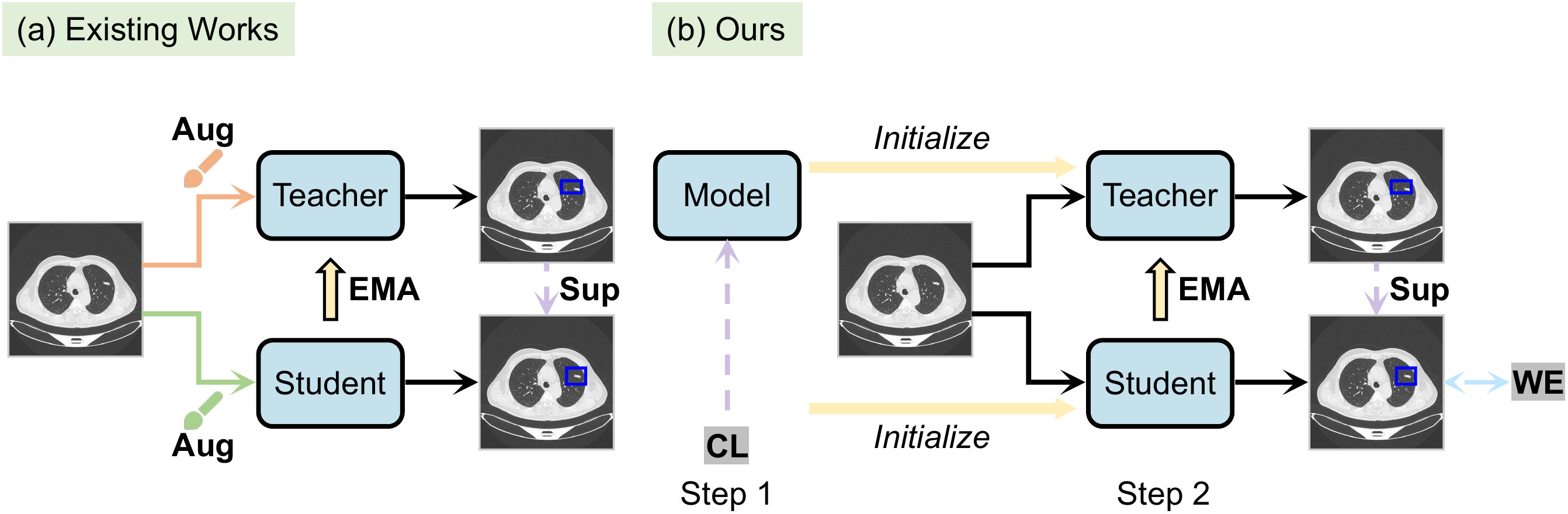}
  \caption{
  (a) Existing SFUDA object detection works utilize feature alignment or sample generation to help with the pseudo labeling. These approaches mainly focus on exploiting the source model. (b) Our proposed SUP-ICI utilizes instance-level contrastive learning (CL) to make use of the foreground-background semantic information of the unlabeled target images. Our weighted entropy (WE) loss is also incorporated for label denoising.
  }
  \label{fig:othervsours}
\end{figure}

However, medical data often involve private information, which makes them not shareable. Consequently, traditional UDA methods, which often rely on access to labeled source data, are not directly applicable in this context. 
Thus in this paper, we aim at the more realistic but challenging source-free unsupervised domain adaptation (SFUDA) setting for pulmonary nodule detection. In this setting, we are confined to utilizing only a pre-trained source model along with unlabeled samples from the target domain, thereby circumventing the need for direct access to the sensitive source data. 
To the best of our knowledge, this is the first work that deals with SFUDA in the pulmonary nodule detection task. Recent SFUDA object detection works utilize the pseudo labeling strategy \cite{aaai21/sfod}, complemented by techniques such as feature alignment or sample generation \cite{nips21/HCL,cvpr22/lods,wacv23/MemCLR}, as illustrated in Fig.~\ref{fig:othervsours} (a), which are consistent with the works in SFUDA image classification and segmentation.

Nonetheless, these works mainly focus on the shifts between the source and target, and neglect the detection's characteristics. For instance, the discrimination of the foreground objects and the backgrounds can naturally be an auxiliary supervision for the target data. Besides, the relatively smaller size of the nodules compared with the objects in natural images degrades the performance of general SFUDA object detection approaches. To address these two limitations for our introduced SFUDA pulmonary nodule detection task, we propose a novel Source-free Unsupervised cross-domain method for Pulmonary nodule detection (SUP), termed Instance-level Contrastive Instruction fine-tuning framework (ICI). The SUP-ICI is a two-step method as illustrated in Fig.~\ref{fig:ours}. First, to leverage the discrepancy between the feature representations of nodules and other entities in the computed tomography (CT) images, we employ instance-level contrastive learning (CL) for adapting the source model to the target domain. This strategy eliminates the requirement for annotations in the target images. More importantly, its use of instance-level \textbf{foreground-background discrimination} realizes to focus on the small-scale features, eliminating being distracted by the dominant backgrounds in pulmonary nodule detection. Given the domain shifts between the source and target domains, we set a high pre-defined threshold for the region proposal network (RPN) classifier to auto-label the nodules for optimal CL. This initial adaptation step enhances the model's ability to accurately detect nodules on the target domain, providing more accurate initial pseudo nodule generation for the second step. Second, the adapted model is duplicated into a teacher model and a student model for further training to elevate the pulmonary nodule detection performance. The teacher functions to generate pseudo nodules for supervising the training of the student. The weights of the student can in turn be utilized to update the teacher to further improve the accuracy of the pseudo nodules. In order to relieve the negative effect of the pseudo label noise, we propose a weighted entropy (WE) loss as an additional unsupervised constraint to facilitate the student training. The WE loss is different from the general entropy loss, and designed particularly for the detection network. It considers that in detection the non-nodule instances are usually much more than the nodules, and re-weights the entropy loss to down-weight the overly confident non-nodule instances and make the detector pay more attention to the less confident nodules. The WE loss also uses intrinsic \textbf{foreground-background discrimination} \cite{mm23/wend} to facilitate the further adaptation of the model.

We summarize our main contributions as follows:
\begin{itemize}
    \item We highlight the setting of SFUDA for pulmonary nodule detection, utilizing only a pre-trained source model and unlabeled target domain images. We curate a benchmark based on four public pulmonary nodule datasets for our proposed new setting.
    \item We propose a novel method termed SUP-ICI for our setting, which can achieve accurate pulmonary nodule detection on the target domain through two steps.
    \item We propose instance-level contrastive learning to adapt the source model to the target domain. 
    By focusing on the disparities between the features of individual instances, our method fosters a deep understanding that enhances the model's adaptation across domains.
    \item We apply the teacher-student framework to enhance detection performance through the generation and refinement of pseudo labels. A specially designed weighted entropy loss further mitigates the impact of inaccuracies in these pseudo labels, making our adaptation process more robust and reliable.
\end{itemize}
Experiments on the benchmark demonstrate the effectiveness of our method, which yields state-of-the-art results.

\section{Related Work}
\label{sec:rws}

\subsection{Pulmonary Nodule Detection}

Pulmonary nodule detection is an effective way of the prevention and treatment of lung cancer. However, manually diagnosing the pulmonary nodules from the hundreds of slices of the CT images is a labor-intensive task. In recent years, with the bloom of the CNN-based methods in various detection tasks \cite{iccv15/fastrcnn,eccv16/ssd}, the pulmonary nodule detection has also been regarded as an object detection task for CT images. Initially, the 2D CNNs are directly utilized for the 3D CT images. In \cite{DBLP:conf/miccai/DingLHW17}, the authors incorporate a deconvolutional structure in Faster RCNN for detecting the nodules on axial slices. In \cite{DBLP:journals/tmi/SetioCLGJRWNSG16}, the multi-view ConvNets is proposed for pulmonary nodule detection, which take a set of 2D patches from differently oriented planes as input. The final results are obtained by delicately fusing the outputs from different streams of 2D ConvNets. 
More recently, to better exploit the 3D spatial information of CT images, many studies focus on using the 3D CNNs, such as \cite{DBLP:journals/tbe/DouCYQH17,deeplung18,DBLP:journals/nn/KimYCS19,leaky_noisy-or19,nodulenet,deepseed20,DBLP:journals/tmi/OzdemirRB20,DBLP:conf/miccai/SongCLHLHCYSZW20,21PAMI-SANet,miccai22/LSSANet,tcbb23/SGDA,DBLP:journals/mia/LuoSWCCLMZ22}. The famous NoduleNet \cite{nodulenet} is an end-to-end 3D deep CNN framework, which achieves the nodule detection, false positive reduction, and nodule segmentation simultaneously in a multi-task manner. Inspired by the success of the Vision Transformers \cite{iclr21/ViT} in natural images, the plug-and-play slice grouped non-local modules \cite{21PAMI-SANet,miccai22/LSSANet} are specially designed for the pulmonary nodule detection task, which can enhance the detector's ability to extract the global information and meanwhile reduce the computation of the original non-local operations \cite{cvpr18/non-local} for the 3D CT images. Although excellent pulmonary nodule detection performance has been achieved under the closed world assumption, these detectors have difficulty in generalizing to the unseen domain \cite{xu2021vitae,zhang2022vitaev2,cvpr/HuangY022}.

\subsection{Unsupervised Domain Adaptation Object Detection}

Unsupervised domain adaptation (UDA) is a practical setting where the labeled source data are provided for adapting to the unlabeled target data. Most existing methods adopt feature alignment for UDA object detection. In \cite{cvpr18/DA_Faster_RCNN}, the authors build image-level and instance-level domain classifiers to implement feature alignment in an adversarial manner. Following this, a strong-weak domain alignment model \cite{cvpr19/Strong_Weak_DA} is proposed to focus on the local and global features separately. Authors of \cite{iccv19/Multi_adversarial_Faster_RCNN} and \cite{iccv19/Multi-level_DA} employ multi-level domain feature alignment. Xu et al. \cite{cvpr20/Cate_regu_DA} propose a categorical regularization framework exploiting the categorical consistency between image-level and instance-level predictions. A center-aware feature alignment method \cite{eccv20/Center_Aware} is developed to allow the discriminator to pay more attention to the foreground features. Some other strategies to deal with foreground and background features are explored \cite{cvpr20/coarse2fine,cvpr21/rpn_align}. In addition to the approaches focusing on the domain shifts, solving the problem of inaccurate label in target domain is another stream \cite{iccv19/Robust_DA,cvpr19/Mean_T_DA,iccv19/Self-training_One,eccv20/tri-way_DA,eccv20/Colla_DA,cvpr21/unbiased_T,cvpr22/Target-P_Distill}. To adapt to the domain shift, the object detector is trained using the source labeled samples and the refined generated annotations in the target domain \cite{iccv19/Robust_DA}. Cai et al. \cite{cvpr19/Mean_T_DA} simulate unsupervised domain adaptation as semi-supervised learning, and integrate the object relations into the measure of consistency cost between teacher and student modules. Cross-domain distillation \cite{cvpr21/unbiased_T} is utilized to alleviate the model bias in Mean-Teacher. He et al. \cite{cvpr22/Target-P_Distill} design a dual-branch self-distillation framework with a cross-domain perceiver for teacher-student mutual learning. Despite their efficacy, all these methods assume access to the source domain, which may cause privacy issues in the medical field. Unlike them, we perform UDA pulmonary nodule detection by only leveraging a pre-trained source model.

\subsection{Source-free Unsupervised Domain Adaptation}

Source-free unsupervised domain adaptation (SFUDA) denotes the setting of adapting to the target domain given only a well-trained source model and unlabeled target data. One stream of the SFUDA methods is implicitly aligning the feature distribution of the source and target domain using the generative adversarial networks (GAN) \cite{MA_cvpr20,VDM-DA_tcsvt22}. Another stream is directly exploiting the knowledge of the source model, especially because the source model can generate noisy labels on the unlabeled target domain \cite{SHOT_icml20,PDALN_EMNLP21,ssnl_iros22}. Qiu et al. \cite{CPGA_ijcai21} propose a two-stage method named CPGA, which first utilizes the classifier of the source model to generate source prototypes via contrastive learning, and then align each pseudo-labeled target data to the corresponding source prototypes. Differently, in \cite{SFDAn_nips21,GSFDA_iccv21}, the authors define the local affinity of the target data, and encourage samples with high local affinity to have consistent predictions. Zhang et al. \cite{DaC_nips22} present a new paradigm called DaC, combining the global class-wise pseudo labeling and the local neighborhood consistency. Besides, some works are tailored for the natural and medical image segmentation \cite{cvpr21/UR_seg,cvpr21/SFDA_seg,mia22/AdaMI,icassp22/APL,mia22/SFDA_fourier}. Despite the impressive progress these approaches have made for the image classification and segmentation tasks, they are not applicable to the detection tasks. They only transfer the semantic information, and are not able to achieve precise instance localization. In addition, there exists an inherent class imbalance in the detection network.

The SFUDA object detection is a relatively new task, and there are not many works in this field. It is first introduced in \cite{aaai21/sfod}, where it is modeled as a noisy label learning problem and solved by using the self-entropy descent policy to search the pseudo label confident threshold. The HCL \cite{nips21/HCL} utilizes historical contrastive learning to exploit the source domain and learn the target representations. LODS \cite{cvpr22/lods} treats the difference between the original target image and the style-enhanced one as an auxiliary supervision for model adaptation. In \cite{wacv23/MemCLR}, the authors introduce a more realistic online SFUDA setting, and design a cross-attention transformer-based memory module for the target representation learning. Nevertheless, these approaches restrict themselves to the image classification and segmentation routines. The characteristics of the object detection are not fully exploited, e.g. the foreground-background discrimination.

\section{Method}
\label{sec:method}

\textbf{Problem Definition.} We aim to address the cross-domain pulmonary nodule detection problem in a source-free unsupervised domain adaptation (SFUDA) setting, where only a pre-trained source model and unlabeled target CT images are available.

\begin{figure*}[!t]
  \centering
  \includegraphics[width=7in]{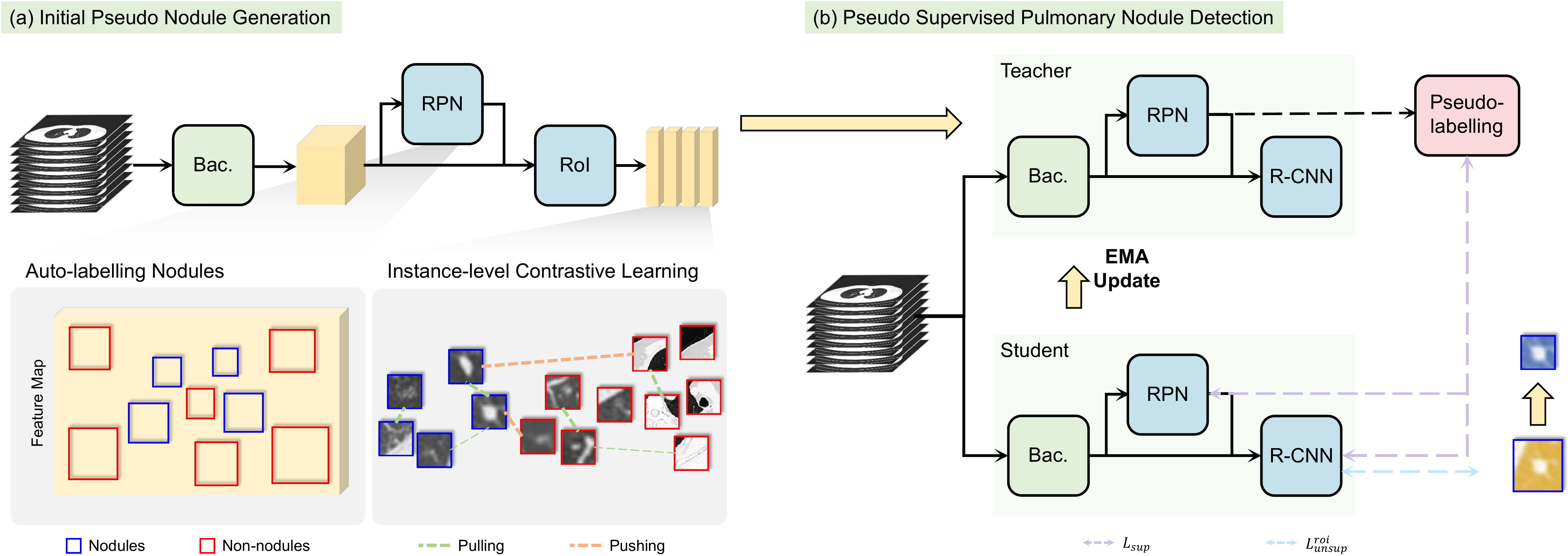}
  \caption{Overview of our Source-free Unsupervised cross-domain method for Pulmonary nodule detection (SUP) named Instance-level Contrastive Instruction fine-tuning framework (ICI). The SUP-ICI consists of two steps: (a) initial pseudo nodule generation and (b) pseudo supervised pulmonary nodule detection. (a) The well-trained source model is adapted to the target domain by instance-level contrastive learning. The proposals of RPN are identified as nodules and non-nodule instances, and cropped from the feature maps as regions of interest (RoI). A positive pair is formed using two nodule or two non-nodule RoI features, whereas a negative pair is formed using one nodule and one non-nodule RoI features. Through contrastive learning, the positive pairs are pulled together and the negative pairs are pushed apart. (b) The adapted model is duplicated into a teacher model and a student model. The teacher is gradually updated by the student using EMA strategy to generate pseudo nodules for the student training. Furthermore, an unsupervised weighted entropy (WE) constraint is added to the R-CNN \cite{iccv15/fastrcnn,nips15/fasterrcnn} to alleviate the negative effect of the pseudo label noise. The WE loss can not only increase the prediction confidence, but also make the localization of the nodule more accurate.}
  \label{fig:ours}
\end{figure*}

\textbf{Overview.} 
Our proposed Source-free Unsupervised cross-domain method for Pulmonary nodule detection (SUP), named Instance-level Contrastive Instruction fine-tuning framework (ICI), contains two steps as shown in Fig.~\ref{fig:ours}. SUP-ICI innovatively combines instance-level contrastive learning and a teacher-student mutual learning framework using weighted entropy loss. 
First, as elaborated in Sec.~\ref{sec:pseudo_label}, the pre-trained source model is adapted to the target domain via instance-level (nodule and non-nodule) contrastive learning for generating accurate pseudo nodules. 
This instance-level \textbf{foreground-background discrimination} not only avoids the requirement of target annotations, but also alleviates the problem of dominance of backgrounds in pulmonary nodule detection, which is common in small-scale object detection tasks. For optimal contrastive learning, we propose an auto-labeling mechanism to select the foreground instances (nodules) and background instances. Second, in Sec.~\ref{sec:nodule_detection}, we duplicate the adapted model into a teacher model and a student model. The teacher model is utilized to generate pseudo nodules for the supervision of the student model training. We choose to also update the teacher model using the student model's weights to improve the accuracy of the pseudo nodules. To mitigate the adverse effect associated with pseudo label noise, we propose a weighted entropy loss as an additional unsupervised constraint for training the student model. This loss facilitates intrinsic \textbf{foreground-background discrimination}, thus aiding the adaptation process. The initial step of our approach ensures the generation of more accurate initial pseudo nodules for the second step, primarily utilizing knowledge from the source domain. The subsequent step then focuses on assimilating and leveraging characteristics unique to the target domain.

\subsection{Initial Pseudo Nodule Generation}
\label{sec:pseudo_label}

\subsubsection{Instance-level Contrastive Learning} To improve the quality of the pseudo nodules generated using the pre-trained source model on the target domain, similar to \cite{cvpr21/ORE,cvpr22/C2AM}, we propose to apply constrastive learning at the instance level, where \underline{nodule and non-nodule} instances form the negative pairs, and \underline{nodule and nodule} instances or \underline{non-nodule and non-nodule} instances form the positive pairs. In other words, the nodule instances and other object instances are pushed apart, and the instances of the same class are pulled together. 

Given $n$ instances identified by the pulmonary nodule detector and their corresponding feature vectors $\textbf{\textit{f}}_{i} \in \mathbb{R}^{d}$, $i=1,2,...,n$, we divide these instances into the foreground nodule set and the background instance set, i.e., $\textbf{\textit{f}}_{1:m}^{+}$ and $\textbf{\textit{f}}_{1:k}^{-}$, where $m + k = n$. Following \cite{cvpr22/C2AM}, 
we form the negative pairs using foreground-background instance pairs, and the positive pairs using the instances from the same sets. The contrastive loss consists of a positive and a negative component, and is defined as follows:

\begin{equation}
\label{eq:ctr_loss}
    L_{ctr} = L_{ctr}^{pos} + L_{ctr}^{neg}.
\end{equation}

The negative contrastive loss is utilized to classify the nodules from other background instances in the latent space:

\begin{align}
    L_{ctr}^{neg} &= - \frac{1}{mk} \sum_{i=1}^m \sum_{j=1}^k \log(1 - s_{i,j}^{neg}), \\
    s_{i,j}^{neg} &= {\rm sim}(\textbf{\textit{f}}_{i}^{+}, \textbf{\textit{f}}_{j}^{-}),
\end{align}

where $s_{i,j}^{neg}$ is the cosine similarity between the feature vectors of each foreground-background instance pair.

For the positive pairs, our goal is to force them to be close. However, the two instances of the positive pairs may have quite different feature representations, since different types of nodules have different morphology and the non-nodule object samples can have any structure. Equally pulling together all positive pairs will cause the positive pairs with large distances impeding the learning process. To alleviate this problem, we propose to reweight each positive contrastive loss using the corresponding positive pair's feature similarity rank among all pairs:

\begin{align}
    L_{ctr}^{pos} &= L_{ctr}^{pos+} + L_{ctr}^{pos-}, \\
    L_{ctr}^{pos+} &= - \frac{1}{m(m-1)} \sum_{i=1}^{m} \sum_{j=1}^{m} \mathbbm{1}_{[i \neq j]} (w_{i,j}^{+} \log (s_{i,j}^{+})), \\
    L_{ctr}^{pos-} &= - \frac{1}{k(k-1)} \sum_{i=1}^{k} \sum_{j=1}^{k} \mathbbm{1}_{[i \neq j]} (w_{i,j}^{-} \log (s_{i,j}^{-})),
\end{align}

where $\mathbbm{1}_{[i \neq j]} \in \{0,1\}$ is an indicator function, whose value is 1 if $i \neq j$. $s_{i,j}^{+}$ or $s_{i,j}^{-}$ denotes the cosine similarity between the feature vectors of each positive pair:

\begin{equation}
    s_{i,j}^{+} = {\rm sim}(\textbf{\textit{f}}_{i}^{+}, \textbf{\textit{f}}_{j}^{+}), s_{i,j}^{-} = {\rm sim}(\textbf{\textit{f}}_{i}^{-}, \textbf{\textit{f}}_{j}^{-}).
\end{equation}

The weight for each positive contrastive loss $w_{i,j}^{+}$ or $w_{i,j}^{-}$ is calculated using the corresponding cosine similarity rank:

\begin{equation}
    w_{i,j}^{+} = {\rm exp}(- \omega {\rm rank}(s_{i,j}^{+})), w_{i,j}^{-} = {\rm exp}(- \omega {\rm rank}(s_{i,j}^{-})),
\end{equation}

where $\omega$ is a smoothing hyper-parameter for modulating the exponential function. ${\rm rank}(s_{i,j}^{+})$ or ${\rm rank}(s_{i,j}^{-})$ represents the rank of cosine similarity $s_{i,j}^{+}$ in the set of foreground nodule pairs $s^{+} = \{s_{1,2}^{+},...,s_{i,j}^{+},...\} (i \neq j)$ or the rank of $s_{i,j}^{-}$ in the set of background instance pairs $s^{-} = \{s_{1,2}^{-},...,s_{i,j}^{-},...\} (i \neq j)$. Through this, the losses of the positive pairs that have similar semantics are assigned large weights, whereas the losses of the positive pairs that are less similar are relatively reduced, thus giving better contrastive learning.

\subsubsection{Auto-labeling Nodules via RPN} While computing the contrastive loss using Eq.~(\ref{eq:ctr_loss}), the instances generated by the pulmonary nodule detector need to be divided into two classes, the nodules and other instances. Due to the significant domain shift between the source and target domains, directly relying on the classification outcomes from the pulmonary nodule detector can lead to unreliable results. To address this issue, we propose to employ a high pre-defined threshold for the region proposal network (RPN) outputs. This ensures that only instances with classification probabilities above this threshold are considered as pseudo nodules, thereby enhancing the quality and reliability of these pseudo labels. Note that we choose to utilize the classification results of RPN to ensure adequate non-nodule instances. Furthermore, the feature vector of each instance is obtained by cropping the corresponding area from the output feature map of the detector's backbone. The incorporation of the auto-labeling mechanism into the instance-level contrastive learning constitutes our first step, as depicted in Fig.~\ref{fig:ours} (a).

\subsection{Pseudo Supervised Pulmonary Nodule Detection}
\label{sec:nodule_detection}

\subsubsection{Teacher-student Mutual Learning} After the preliminary domain adaptation through contrastive learning, the adapted model is able to detect reliable nodules on the target domain. To further improve the pulmonary nodule detection accuracy, we apply the teacher-student mutual learning framework as shown in Fig.~\ref{fig:ours} (b), where a teacher model and a student model are first initialized using the weights of the adapted model $\theta$, i.e. $\theta_{t} \leftarrow \theta,\theta_{s} \leftarrow \theta$, and then iteratively refine each other's performance. Initially, both models start with the same weights. Then during training, the student model is optimized using the pseudo nodules generated by the teacher model; specifically, the teacher model's RPN outputs are utilized for jointly supervising the student model's RPN and Region-based CNN (R-CNN). Meanwhile, the teacher model is also gradually updated using the student's weights and the exponential moving average (EMA) strategy. This iterative refinement enhances the detection accuracy by continuously improving the quality of pseudo nodule annotations and the models' ability to detect nodules accurately.

For the student model training, to avoid the accumulative errors caused by the noisy pseudo labels, we set a confidence threshold $\delta$ for the classification scores of the teacher's RPN predictions to filter low-confident predicted nodules, which are mostly false positives. This threshold is critical for minimizing noise in the training data and ensuring that the student model learns from reliable annotations. 
Then the retained nodules are treated as pseudo ground truths for training the student model, to supervise the classification and regression predictions from both RPN and R-CNN; namely, the student model's weights $\theta_{s}$ are optimized using the following pseudo supervised loss:

\begin{align}
    L_{sup} &= L_{sup}^{rpn} + L_{sup}^{roi}, \\ 
    L_{sup}^{rpn} &= \sum_{i} L_{cls}^{rpn}(p_{i},p_{i}^{*}) + L_{reg}^{rpn}(t_{i},t_{i}^{*}), \\
    L_{sup}^{roi} &= \sum_{j} L_{cls}^{roi}(p_{j},p_{j}^{*}) + L_{reg}^{roi}(t_{j},t_{j}^{*}),
\end{align}

where $L_{sup}^{rpn}$ is the loss for predictions from RPN and $L_{sup}^{roi}$ is the loss for the refined predictions from R-CNN. They both consist of a classification loss and a regression loss. $p_{i}$ and $p_{j}$ denote the predicted probability of the $i$-th instance being a nodule from RPN and the predicted probability of the $j$-th instance from R-CNN, respectively. $t_{i}$ and $t_{j}$ are vectors denoting the 6 parameterized coordinates of the predicted nodule. The asterisked variables, $p_{i}^{*}$, $p_{j}^{*}$, $t_{i}^{*}$, and $t_{j}^{*}$, denote their corresponding pseudo ground-truth counterparts, which are assigned using the teacher's PRN outputs.

To enable stable pseudo nodule generation, the teacher model is gradually updated by the student model's weights using the EMA strategy. Thus it can be regarded as multiple temporal ensembled student models. For each time step, the teacher model weights are updated as:

\begin{equation}
    \theta_{t} \leftarrow \beta \theta_{t} + (1 - \beta) \theta_{s},
\end{equation}

where $\beta \in [0,1]$ is a smoothing hyper-parameter, used to modulate the contribution of the historical weights of the teacher model to its update process. By blending the teacher's historical weights with the student's current weights, EMA ensures that the teacher model evolves smoothly, avoiding abrupt changes that could introduce instability. This strategy helps maintain a balance between leveraging the student model's latest insights and preserving the teacher model's accumulated knowledge.

\subsubsection{Weighted Entropy Loss} In order to further reduce the negative impact of the pseudo nodule noise, we propose to introduce an additional unsupervised constraint for the student model training and design a weighted entropy (WE) loss. Considering the success of the entropy loss in dealing with the unlabeled data in the semi-supervised and unsupervised image classification tasks, we employ the entropy loss as an additional unsupervised constraint for the nodule and non-nodule classification on the unlabeled target domain. For one thing, the object recognition is much more difficult than the localization as claimed in \cite{arxiv/classify_hardthan_detect}, and for another, we assume that the confident classification enforces the accurate localization. However, different from the general image classification, there exists a class imbalance issue in object detection. In pulmonary nodule detection, the predictions of the non-nodule instances are dominant and overly confident, whereas the predictions of the nodules are the opposite. Simply adopting the original entropy loss can increase this nodule and non-nodule imbalance problem. Similar to the Focal loss\cite{iccv17/focalloss}, we propose to re-weight the original entropy loss using the predicted probability, thus down-weighting the easy instances and focusing on the hard ones for the classifier of the R-CNN \cite{iccv15/fastrcnn,nips15/fasterrcnn}:

\begin{gather}
  L_{unsup}^{roi} = \sum_{j}H_{cls}^{roi}(p_{j}, \tau_{1}, \tau_{2}), \\
  \begin{aligned}
  H_{cls}^{roi}(p_{j}, \tau_{1}, \tau_{2}) = - (\mathbbm{1}(p_{j} < \tau_{1})(1 - \alpha)p_{j}^{\gamma} \\ + \mathbbm{1}(p_{j} > \tau_{2})\alpha(1 - p_{j})^{\gamma})p_{j}\log(p_{j}),
  \end{aligned}
  \label{unsup_loss}
\end{gather}

where $\tau_{1}$ and $\tau_{2}$ are the confidence thresholds for the choice of the modulating factor $p_{j}^{\gamma}$ or $(1 - p_{j})^{\gamma}$ for the original entropy $p_{j}\log(p_{j})$.
Both the indicator and modulating factor are solely related to the predicted probability rather than the generated pseudo nodule labels. Thus they are determined by the intrinsic foreground-background discrimination instead of the model predictions. Then the selected modulating factor is able to down-weight the well-classified instances' entropy and focus on the hard ones. As in \cite{iccv17/focalloss}, $\gamma$ is a tunable focusing parameter to control the rate of down-weighting, and $\alpha$ is another hyper-parameter for dealing with the class imbalance. It is noteworthy that our proposed modulating factor choice mechanism here can be treated as a self-paced learning method \cite{ijcai15/SPL,ijcai21/SAIL}. Since the instances which have predicted probabilities between $\tau_{1}$ and $\tau_{2}$ are neglected, at the beginning of the second step, there are not much instances involved in calculating the WE loss, and as the training progresses, more and more instances are used. This further ensures the modulating factor being proper.

To sum up, the loss function for training the student model comprises the supervised loss given the pseudo ground-truth nodules generated by the teacher model and the unsupervised WE loss:
\begin{equation}
  L_{s} = \eta L_{sup} + L_{unsup}^{roi},
  \label{total_loss}
\end{equation}
where $\eta \geq 0$ is a balancing hyper-parameter.

\section{Experiments}
\label{sec:exp}

\subsection{Experimental Setups}
\subsubsection{Datasets} 
\begin{figure}[!t]
\centering
\includegraphics[width=3.2in]{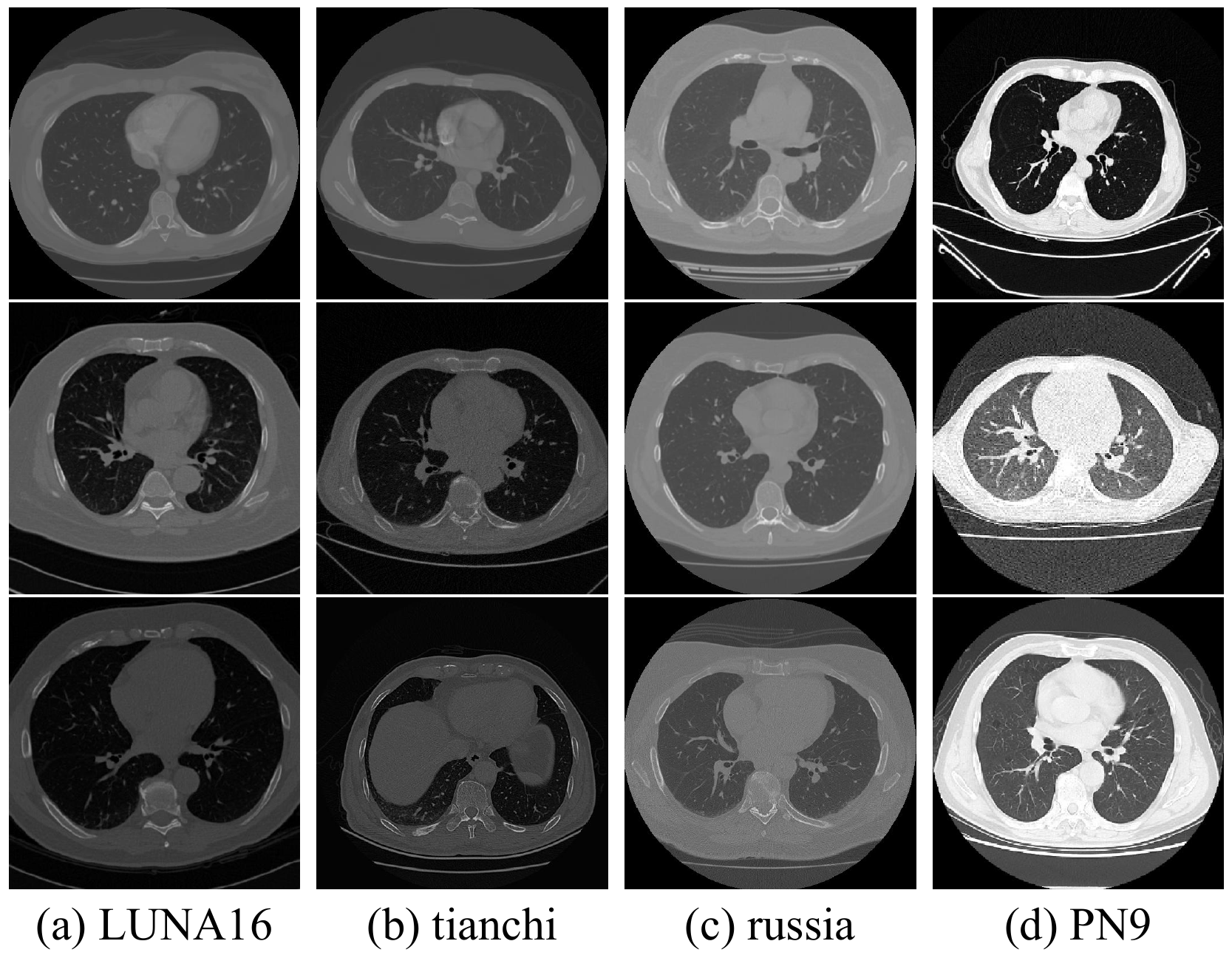}
\caption{Samples of four pulmonary nodule datasets. Each column belongs to one pulmonary nodule dataset as labeled. CT images of different datasets present domain discrepancy, such as illumination, color contrast/saturation, resolution, number of nodules.}
\label{dataset}
\end{figure}

We benchmark SFUDA pulmonary nodule detection by adapting the model pre-trained on PN9 \cite{21PAMI-SANet} to three target datasets: LUNA16 \cite{luna16}, tianchi \cite{tianchi}, and russia \cite{russia}. Samples in Fig.~\ref{dataset} show domain shifts between datasets. The source dataset PN9 \cite{21PAMI-SANet} is so far the largest and most diverse pulmonary nodule dataset collected from various sites. 
Notably, in our setting, we do not have access to the source dataset PN9 \cite{21PAMI-SANet}; here we present it to better illustrate the domain shift. The target LUNA16 \cite{luna16} is a widely used public dataset for pulmonary nodule detection from the LUNA16 challenge. Similarly, the target tianchi \cite{tianchi} is an open dataset from the tianchi medical AI competition. The target russia \cite{russia} is a relatively challenging dataset for its limited quantity of CT images and large amounts of nodules. Each CT image in the target datasets is labeled one or more nodules. In our work, the annotations are only utilized for evaluation. Following \cite{tcbb23/SGDA}, we utilize subsets comprising 601, 800, and 364 CT images from the LUNA16 \cite{luna16}, tianchi \cite{tianchi}, and russia \cite{russia}, respectively. They are partitioned into the commonly used ratios of 7:1:2 for training, validation, and testing, respectively. 

\subsubsection{Evaluation Metric} We use the Free-Response Receiver Operating Characteristic (FROC) for evaluation. It is the average of recall rates at 0.125, 0.25, 0.5, 1, 2, 4, and 8 false positives per CT image. A predicted nodule is correct if it is located within a distance $R$ from the center of any annotated nodules, where $R$ is the radius of the annotated nodule. Predicted nodules not located in the range of any annotated nodules are seen as false positives.

\subsubsection{Data Preprocessing} 
Following the regular procedure in the previous works, we adopt three steps to preprocess the raw CT data in the three datasets. First, in order to reduce the unnecessary calculations, we employ the lungmask \cite{lungmask} to segment the lung regions from the CT images. The Hounsfield Unit (HU) values of the raw CT data are clipped into $[-1200,600]$, and transformed linearly into $[0,255]$ to obtain unit8 values. Then the values of the voxels not in the lung regions are assigned a padding value of 170. Second, all the CT images are resampled to $1\times1\times1$ mm spacing. Third, in order to further decrease the calculation load, we crop the CT images using the segmented lung boxes to reduce the image sizes. For convenience, we adopt the voxel coordinates during the training and inference. The annotation coordinates are modified according to the aforementioned preprocessing processes for evaluation.

\subsubsection{Patch-based Input} In the training phase, considering the high computational complexity of the 3D CNN and the limitation of the GPU memory, using the entire CT images as input is not feasible. Thus, the CT images are cropped into small 3D patches with the size of $1\times128\times128\times128$ (Channel$\times$Depth$\times$Height$\times$Width), and each 3D patch is treated as an individual input. Some patches may exceed the range of the CT images from which they are extracted; the outside part is filled with a value of 170. In the test phase, the full CT images are fed into the well-trained detector without being cropped. The CT images are padded with a value of 170 to ensure that no odd size of input for the detector exists.

\subsubsection{Implementation Details} Since our setting assumes a given source pre-trained model, we adopt the SANet \cite{21PAMI-SANet} as the pulmonary nodule detector, thus to directly use the weights pre-trained on PN9 provided in \cite{21PAMI-SANet}. With regards to the model training on the target, we use the Stochastic Gradient Descent (SGD) optimizer with a batch size of 8. The learning rate is set to be 0.0005; the momentum and weight decay coefficients are set to be 0.9 and $1\times10^{-4}$, respectively. The maximum number of training epochs for each stage is set as 100. Other hyper-parameters are kept the same with those in \cite{21PAMI-SANet}. All the experiments are implemented using PyTorch on NVIDIA GeForce RTX 3090 GPUs with 24GB memory.

\begin{figure*}[!t]
  \centering
  \includegraphics[width=7in]{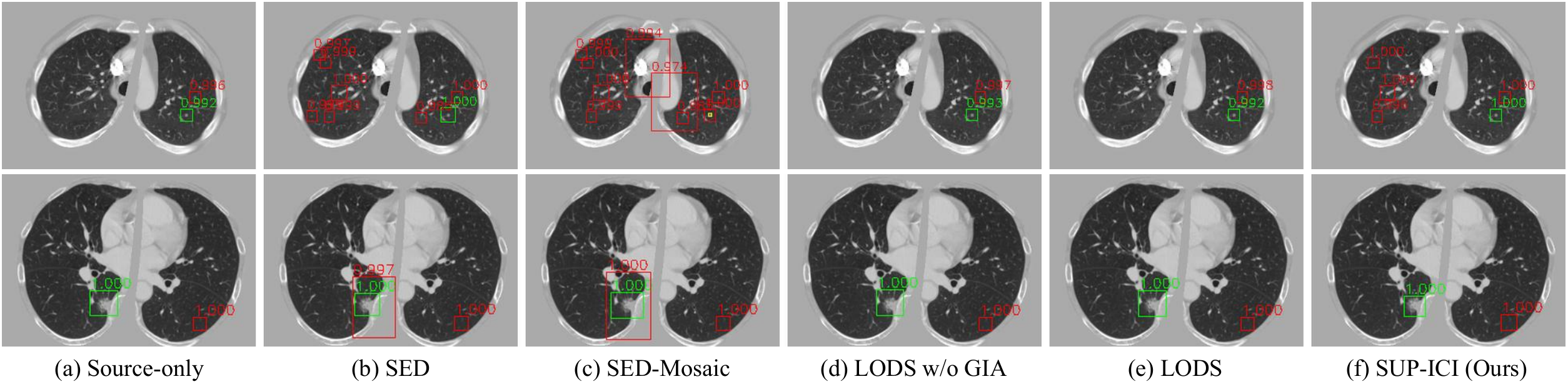}
  \caption{Exemplar detection results of the compared approaches and our method SUP-ICI. The green boxes, red boxes, and yellow boxes denote the true positives (TP), false positives (FP), and false negatives (FN), respectively. The values above the boxes are the detection scores.}
  \label{fig:visual}
\end{figure*}

\subsection{Comparison with the State-of-the-arts}
We compare our SUP-ICI with several state-of-the-art (SOTA) SFUDA detection approaches including the SED \cite{aaai21/sfod} and LODS \cite{cvpr22/lods} on our established benchmark scenarios: PN9 $\rightarrow$ LUNA16, PN9 $\rightarrow$ tianchi, and PN9 $\rightarrow$ russia. We re-implement the compared approaches on our benchmark scenarios using the same base detector SANet \cite{21PAMI-SANet} as our SUP-ICI. `Source-only' and `Target-supervised' denote the SANet \cite{21PAMI-SANet} trained using the labeled source data and target data respectively. The experiment results are listed in Table~\ref{tab:luna16}, Table~\ref{tab:tianchi}, and Table~\ref{tab:russia}. As shown, our method achieves the best results in terms of the FROC score and outperforms the other approaches by a large margin. On the scenario PN9 $\rightarrow$ russia, the performance of our SUP-ICI even surpasses the Target-supervised approach.

\begin{table}[!t]
\caption{Lung nodule detection sensitivities (\%) on PN9 $\rightarrow$ LUNA16. Each column denotes the sensitivities corresponding to different average number of false positives per CT image.}
\label{tab:luna16}
\centering
\setlength{\tabcolsep}{0.75mm}
\begin{tabular}{l | c c c c c c c c} \hline
Method                  & 0.125 & 0.25  & 0.5   & 1     & 2     & 4     & 8     & Avg   \\ \hline
Source-only             & 26.14 & 37.15 & \underline{53.66} & 61.46 & \textbf{75.22} & \underline{82.56} & 88.07 & 60.61 \\
SED \cite{aaai21/sfod} & \underline{49.54} & \underline{49.54} & 49.54 & 49.54 & 55.50 & 68.80 & 78.44 & 57.27 \\
SED-Mosaic \cite{aaai21/sfod} & 44.49 & 44.49 & 44.49 & 52.75 & 59.63 & 68.80 & 79.35 & 56.29 \\
LODS w/o GIA \cite{cvpr22/lods} & 25.22 & 38.07 & 51.83 & \underline{66.51} & 74.31 & \textbf{83.48} & \textbf{91.28} & \underline{61.53} \\
LODS \cite{cvpr22/lods} & 23.39 & 35.32 & 47.70 & 61.92 & 72.01 & 80.73 & \underline{89.90} & 58.71 \\
SUP-ICI (Ours)              & \textbf{74.77} & \textbf{74.77} & \textbf{74.77} & \textbf{74.77} & \underline{74.77} & 81.65 & 87.61 & \textbf{77.58} \\ \hline
Target-supervised       & 57.33 & 66.97 & 74.77 & 82.11 & 87.61 & 92.66 & 94.49 & 79.42 \\ \hline
\end{tabular}
\end{table}

\begin{table}[!t]
\caption{Lung nodule detection sensitivities (\%) on PN9 $\rightarrow$ tianchi.}
\label{tab:tianchi}
\centering
\setlength{\tabcolsep}{0.75mm}
\begin{tabular}{l | c c c c c c c c} \hline
Method                  & 0.125 & 0.25  & 0.5   & 1     & 2     & 4     & 8     & Avg   \\ \hline
Source-only             & 20.52 & 35.44 & 48.88 & 61.19 & \underline{77.23} & 85.44 & 91.04 & 59.96 \\
SED \cite{aaai21/sfod} & 51.49 & 51.49 & 51.49 & 51.49 & 60.82 & 67.16 & 79.85 & 59.11 \\
SED-Mosaic \cite{aaai21/sfod} & \underline{52.61} & \underline{52.61} & \underline{52.61} & 58.58 & 68.28 & 76.86 & 85.82 & \underline{63.91} \\
LODS w/o GIA \cite{cvpr22/lods} & 27.98 & 35.44 & 48.13 & \underline{64.55} & \textbf{77.98} & \textbf{89.92} & \textbf{92.91} & 62.42 \\
LODS \cite{cvpr22/lods} & 20.89 & 32.46 & 47.01 & 63.80 & 75.37 & \underline{89.17} & \underline{92.53} & 60.18 \\
SUP-ICI (Ours)              & \textbf{72.01} & \textbf{72.01} & \textbf{72.01} & \textbf{72.01} & 72.01 & 79.10 & 84.70 & \textbf{74.84} \\ \hline
Target-supervised       & 54.85 & 65.29 & 73.88 & 80.97 & 86.56 & 93.65 & 96.26 & 78.78 \\ \hline
\end{tabular}
\end{table}

\begin{table}[!t]
\caption{Lung nodule detection sensitivities (\%) on PN9 $\rightarrow$ russia.}
\label{tab:russia}
\centering
\setlength{\tabcolsep}{0.75mm}
\begin{tabular}{l | c c c c c c c c} \hline
Method                  & 0.125 & 0.25  & 0.5   & 1     & 2     & 4     & 8     & Avg   \\ \hline
Source-only             & 3.04 & 4.70 & 8.86 & 18.00 & 25.20 & 40.16 & 49.86 & 21.40 \\
SED \cite{aaai21/sfod} & \underline{26.59} & \underline{26.59} & \underline{26.59} & \underline{26.59} & 26.59 & 31.02 & 41.27 & \underline{29.32} \\
SED-Mosaic \cite{aaai21/sfod} & 21.88 & 21.88 & 21.88 & 21.88 & 21.88 & 25.76 & 35.45 & 24.37 \\
LODS w/o GIA \cite{cvpr22/lods} & 3.32 & 4.43 & 6.92 & 18.28 & \underline{27.70} & \textbf{42.65} & \textbf{54.01} & 22.47 \\
LODS \cite{cvpr22/lods} & 2.49 & 3.60 & 6.37 & 16.62 & 27.14 & \underline{40.72} & \underline{52.90} & 21.40 \\
SUP-ICI (Ours)              & \textbf{36.84} & \textbf{36.84} & \textbf{36.84} & \textbf{36.84} & \textbf{36.84} & 36.84 & 47.09 & \textbf{38.30} \\ \hline
Target-supervised       & 11.35 & 16.89 & 24.93 & 34.34 & 47.36 & 58.72 & 72.57 & 38.02 \\ \hline
\end{tabular}
\end{table}

It is noted that the SED \cite{aaai21/sfod} even performs worse than the baseline approach Source-only on the scenarios PN9 $\rightarrow$ LUNA16 and PN9 $\rightarrow$ tianchi. The recall rates at the small average numbers of false positives per CT image are better than the baseline whereas the recall rates at the large ones are worse than the baseline. This indicates that although the SED \cite{aaai21/sfod} improves the detection performance of the easily detected nodules, it has much difficulty in detecting the hard nodules, which is consistent with the insight in \cite{aaai21/sfod}. Besides, different from the object detection in natural images, there usually exist many false positives in the pulmonary nodule detection task, causing more and more pseudo nodules in the SED \cite{aaai21/sfod} approach. This further hinders the discrimination of the nodules from other tissues and degrades the performance of the SED \cite{aaai21/sfod}. The higher FROC score of the SED \cite{aaai21/sfod} than the baseline on the scenario PN9 $\rightarrow$ russia is due to the higher detection recall rates at the small average numbers of false positives per CT image. When the Mosaic augmentation \cite{arxiv20/yolov4} is added to mine the false negatives, the FROC score of the SED \cite{aaai21/sfod} improves on the PN9 $\rightarrow$ tianchi but drops on the PN9 $\rightarrow$ LUNA16 and PN9 $\rightarrow$ russia. This reveals that the Mosaic augmentation \cite{arxiv20/yolov4}, which works well in the natural images, may not be applicable to the small-sized lung nodule detection. We also believe that the characteristics of the SED \cite{aaai21/sfod} are not fully applicable to this binary-class detection task.

\begin{table*}[!t]
\caption{Ablation study for each constituent component of our method on PN9 $\rightarrow$ LUNA16 (\%). `CL' refers to the instance-level contrastive learning, `PS' means the pseudo supervised training without updating the teacher, `T-S' denotes the interactive teacher-student framework, and `WE' represents the proposed weighted entropy loss.}
\label{tab:ab_parts}
\centering
\setlength{\tabcolsep}{3.65mm}
\begin{tabular}{l | c c c c | c c c c c c c c} \hline
Method                  & CL & PS & T-S & WE & 0.125 & 0.25  & 0.5   & 1     & 2     & 4     & 8     & Avg   \\ \hline
Source-only             &                      &                   &            & 
           & 26.14 & 37.15 & 53.66 & 61.46 & 75.22 & 82.56 & 88.07 & 60.61 \\
(a)              & \checkmark           &                   &            & 
           & 35.77 & 47.24 & 59.63 & 68.80 & \textbf{76.14} & \textbf{85.32} & \textbf{92.66} & 66.51 \\
(b1)             &                      &  \checkmark       &            &
           & 51.83 & 51.83 & 55.50 & 65.59 & 72.93 & 83.94 & 91.28 & 67.56 \\
(b2)             & \checkmark           &  \checkmark       &            & 
           & 67.43 & 67.43 & 67.43 & 67.43 & 75.22 & \textbf{85.32} & 90.82 & 74.44 \\
(b3)             &                      &  \checkmark       &            &
\checkmark & 47.24 & 47.24 & 53.21 & 61.46 & 71.10 & 82.56 & 89.90 & 64.67 \\
(b4)             &  \checkmark          &   \checkmark      &            &
\checkmark & 66.51 & 66.51 & 66.51 & 66.51 & 72.01 & 82.11 & 87.61 & 72.54 \\
(c1)             &                      &                   & \checkmark &
           & 34.86 & 40.82 & 50.45 & 62.84 & 71.55 & 83.48 & 89.33 & 61.92\\
(c2)             & \checkmark           &                   & \checkmark & 
           & 72.47 & 72.47 & 72.47 & 72.47 & 72.47 & 79.35 & 85.77 & 75.36 \\
(c3)             &                      &                   & \checkmark & 
\checkmark & 31.19 & 36.69 & 49.54 & 62.84 & 71.55 & 83.94 & 89.90 & 60.81 \\
(c4)             & \checkmark           &                   & \checkmark & 
\checkmark & \textbf{74.77} & \textbf{74.77} & \textbf{74.77} & \textbf{74.77} & 74.77 & 81.65 & 87.61 & \textbf{77.58}  \\ \hline
Target-supervised       &                      &                   &            & 
           & 57.33 & 66.97 & 74.77 & 82.11 & 87.61 & 92.66 & 94.49 & 79.42 \\ \hline
\end{tabular}
\end{table*}

On the contrary, the LODS \cite{cvpr22/lods} is capable of detecting more nodules as can be seen from the recall rates at the large average numbers of false positives per CT image. Nonetheless, its ability to distinguish the nodules from other tissues is weaker. It should be noted that our re-implementation of the LODS \cite{cvpr22/lods} removes the style enhancement module. This is because the reconstruction of the CT images itself is already a challenging problem. Then without the style-enhanced target images, the overlooking style module turns into a basic Mean-Teacher framework. Since the graph-based class-wise instance-level alignment in the overlooking style module is only applied to the positive samples and the pulmonary nodule detection is a binary-class detection task, it is applied to a single class in the pulmonary nodule detection. Thus the discriminative features of the nodules are not fully explored and the classification of the nodules is not satisfactory. Moreover, after adding the graph-based image-level alignment, the performance of the LODS \cite{cvpr22/lods} even decreases. We believe the reason is that, in small-sized pulmonary nodule detection, the image-level features involve much more voxel information than the instance-level features and its alignment causes disturbance to the instance-level alignment. These reveal that the SOTA SFUDA object detection approaches for natural images usually either cannot be utilized directly or cannot achieve the satisfactory results in the medical field.

\textbf{Qualitative Comparison.} We visualize the detection results of the compared SOTA approaches and our SUP-ICI in Fig.~\ref{fig:visual}. The detection results are displayed based on the central slices of the nodule ground truths. The true positives (TP), false positives (FP), and false negatives (FN) are denoted by the green boxes, red boxes, and yellow boxes, respectively. We also display the detection scores above the corresponding boxes. As shown, compared with other approaches, our SUP-ICI is able to detect the TPs with higher confidences and more precise positions. On the contrary, the compared approaches' detection results are usually offset or even completely miss the ground truths. Besides, some FPs of the compared approaches have higher detection scores than the TPs, which degrades the performance of these approaches. Specifically, the SED \cite{aaai21/sfod} tends to miss the hard nodules, and the LODS \cite{cvpr22/lods} tends to predict higher scores for the FPs, which are consistent with the quantitative analyses. These results demonstrate the superiority of our SUP-ICI on the source-free unsupervised cross-domain pulmonary nodule detection task.

\subsection{Ablation Study}

We first analyze the effectiveness of each component of our SUP-ICI by isolating each of them on the LUNA16 \cite{luna16}. The results are displayed in Table~\ref{tab:ab_parts}. We choose the model pre-trained on the source dataset PN9 \cite{21PAMI-SANet} as our baseline and directly test it on the target dataset LUNA16 \cite{luna16}. The methods (a), (b1), and (c1) add the instance-level contrastive learning (CL), simple pseudo supervised training without updating the teacher (PS), and teacher-student mutual training (T-S) to the baseline separately. The unlabeled target data are utilized for their training. As the results shown in Table~\ref{tab:ab_parts}, each part contributes to the pulmonary nodule detection. Thereinto, the PS obtains the largest performance gain of more than 7\%. Method (b2) applies the combination of the CL and PS; method (c2) applies the combination of the CL and T-S. From the results in Table~\ref{tab:ab_parts}, we can see that using more accurate pseudo nodules generated by the CL (66.51\% vs. 60.61\%), the PS and T-S work better, and improve the baseline by 13.83\% and 14.75\% respectively. In this circumstance (w/ CL), the superiority of T-S over PS is highlighted. Method (b4) and (c4) are our complete SUP-ICI, and method (b3) and (c3) removes the CL of our SUP-ICI. As shown in Table~\ref{tab:ab_parts}, our SUP-ICI (c4) achieves the best performance, which outperforms other methods (a-c4) by a large margin and improves the baseline by approximately 17\%. It is noted that method (b3)/(c3) is not only surpassed by the SUP-ICI (b4)/(c4) but also by the method (b1)/(c1). This reveals that the WE loss is more useful when the pseudo labels are reliable. 

\begin{table}[!t]
\caption{Ablation study for the teacher-student framework using different threshold $\delta$ to filter the pseudo nodules on PN9 $\rightarrow$ LUNA16 (\%). The teacher model is not updated and the WE loss is not applied.}
\label{tab:ab_delta}
\centering
\setlength{\tabcolsep}{1.84mm}
\begin{tabular}{c | c c c c c c c c} \hline
$\delta$                & 0.125 & 0.25  & 0.5   & 1     & 2     & 4     & 8     & Avg   \\ \hline
0.95                    & 45.41 & 54.12 & 61.92 & 69.72 & \textbf{75.22} & \textbf{86.69} & \textbf{93.11} & 69.46 \\
0.9                     & 60.09 & 60.09 & 60.09 & 65.59 & 73.85 & 83.02 & 89.90 & 70.38 \\
0.8                     & 63.30 & 63.30 & 63.30 & 68.34 & 73.39 & 81.19 & 90.82 & 71.95 \\ 
0.7                     & 67.43 & 67.43 & 67.43 & 67.43 & \textbf{75.22} & 85.32 & 90.82 & 74.44 \\
0.6                     & 65.59 & 65.59 & 65.59 & 69.26 & 72.01 & 83.48 & 88.53 & 72.87 \\
0.5                     & \textbf{72.01} & \textbf{72.01} & \textbf{72.01} & \textbf{72.01} & 74.77 & 82.11 & 86.23 & \textbf{75.88} \\ \hline
\end{tabular}
\end{table}

Secondly, we take a deep look into the applied T-S framework. We start with the study of different confidence threshold $\delta$, which is used for filtering the low-confidence predicted pseudo nodules. In this process, we do not update the teacher model, i.e. PS, and the results are listed in Table~\ref{tab:ab_delta}. As seen, the $\delta$ value of 0.7 yields the best overall performance, which is consistent with our expectation. Intuitively, higher $\delta$ gives more precise pseudo nodules but the number of nodules is limited for learning the discriminative nodule features. On the contrary, lower $\delta$ gives more nodules but at the same time brings more noise in the labels, biasing the model towards the easily detected nodules. Thus the model using higher $\delta$ cannot achieve satisfactory results, and neither does the model using lower $\delta$.

\begin{table}[!t]
\caption{Ablation study for the teacher-student framework using different EMA rate $\beta$ to update the teacher model on PN9 $\rightarrow$ LUNA16 (\%). $\delta$ is set to be 0.7 and the WE loss is not applied.}
\label{tab:ab_beta}
\centering
\setlength{\tabcolsep}{1.55mm}
\begin{tabular}{c | c c c c c c c c} \hline
$\beta$                 & 0.125 & 0.25  & 0.5   & 1     & 2     & 4     & 8     & Avg   \\ \hline
0.999                   & \textbf{74.77} & \textbf{74.77} & \textbf{74.77} & \textbf{74.77} & 74.77 & 77.06 & 82.11 & \textbf{76.14} \\
0.9996                  & 72.47 & 72.47 & 72.47 & 72.47 & 72.47 & \textbf{79.35} & 85.77 & 75.36 \\ 
0.9999                  & 35.77 & 48.62 & 59.63 & 69.26 & \textbf{76.60} & 86.23 & \textbf{93.11} & 67.03 \\
0.99999                 & 36.23 & 47.24 & 59.17 & 68.80 & \textbf{76.60} & 85.77 & \textbf{93.11} & 66.71 \\ \hline
\end{tabular}
\end{table}

\begin{table}[!t]
\caption{Ablation study for the proposed WE loss using different $\gamma$ and $\alpha$ on PN9 $\rightarrow$ LUNA16 (\%). $\delta$ and $\beta$ are set to be 0.7 and 0.9996 respectively. $\eta$ is set to be 1.}
\label{tab:ab_WE}
\centering
\setlength{\tabcolsep}{1.55mm}
\begin{tabular}{c c | c c c c c c c c} \hline
$\gamma$ & $\alpha$ & 0.125 & 0.25  & 0.5   & 1     & 2     & 4     & 8     & Avg   \\ \hline
2        & 0.5      & 74.31 & 74.31 & 74.31 & 74.31 & 74.31 & 78.89 & 83.02 & 76.21 \\
2        & 0.25     & 75.22 & 75.22 & 75.22 & 75.22 & 75.22 & 79.81 & 84.40 & 77.19 \\
4        & 0.25     & \textbf{75.68} & \textbf{75.68} & \textbf{75.68} & \textbf{75.68} & \textbf{75.68} & 77.06 & 83.02 & 76.93 \\
4        & 0.1      & 74.77 & 74.77 & 74.77 & 74.77 & 74.77 & \textbf{81.65} & \textbf{87.61} & \textbf{77.58} \\ 
6        & 0.1      & 72.47 & 72.47 & 72.47 & 72.47 & 71.47 & 77.98 & 83.94 & 74.90 \\ 
6        & 0.05     & 73.39 & 73.39 & 73.39 & 73.39 & 73.39 & 73.39 & 78.44 & 74.11 \\ \hline
\end{tabular}
\end{table}

\begin{table}[!t]
\caption{Ablation study for the proposed method using different $\eta$ on PN9 $\rightarrow$ LUNA16 (\%). $\delta$ and $\beta$ are set to be 0.7 and 0.9996 respectively. $\gamma$ and $\alpha$ are set to be 4 and 0.1 respectively.}
\label{tab:ab_eta}
\centering
\setlength{\tabcolsep}{1.7mm}
\begin{tabular}{c | c c c c c c c c} \hline
$\eta$                  & 0.125 & 0.25  & 0.5   & 1     & 2     & 4     & 8     & Avg   \\ \hline
4                       & \textbf{74.77} & \textbf{74.77} & \textbf{74.77} & \textbf{74.77} & 74.77 & 80.73 & 86.69 & 77.32  \\
2                       & 72.47 & 72.47 & 72.47 & 72.47 & 72.47 & 77.06 & 83.94 & 74.77 \\
1                       & \textbf{74.77} & \textbf{74.77} & \textbf{74.77} & \textbf{74.77} & 74.77 & 81.65 & 87.61 & \textbf{77.58} \\ 
0.5                     & 69.26 & 69.26 & 69.26 & 69.26 & 75.22 & 83.48 & 88.99 & 74.96 \\ 
0.25                    & 71.55 & 71.55 & 71.55 & 71.55 & 71.55 & 81.65 & 85.32 & 74.96  \\
0.125                   & 66.51 & 66.51 & 66.51 & 67.88 & 72.47 & 81.19 & 87.15 & 72.60 \\
0.0625                  & 55.04 & 55.04 & 64.22 & 69.72 & \textbf{76.14} & \textbf{84.86} & \textbf{90.36} & 70.77 \\ \hline
\end{tabular}
\end{table}

We also analyze the impact of various EMA rate $\beta$ on the model performance. Smaller $\beta$ means more student model's weights are transferred to the teacher model, which is likely to bring more noise in the pseudo labels and especially cause false negatives. This can be seen from the recall rates at the large average numbers of false positives per CT image in Table~\ref{tab:ab_beta}. However, if the $\beta$ is too large, then the teacher model will barely change. These results demonstrate the necessity of dealing with the label noise for the teacher-student mutual learning.

Thirdly, we investigate our proposed WE loss. There are two hyper-parameters in the WE loss, namely $\gamma$ and $\alpha$. We choose to alter these two parameters together for analysis because the focusing parameter $\gamma$ down-weights mainly the easy non-nodules, as the balancing parameter $\alpha$ is employed to trade this off. The results of combination of different $\gamma$ and $\alpha$ are shown in Table~\ref{tab:ab_WE}. It can be observed that our WE loss is beneficial to the model performance, and proper choices of $\gamma$ and $\alpha$ in the WE loss increase the detection performance up to 2.22\%.

Finally, we study the influence of balancing the weights between the pseudo supervised and unsupervised loss during T-S training by varying the balancing parameter $\eta$. As shown in Table~\ref{tab:ab_eta}, the results of $\eta$ 1 are the best. 

\section{Conclusion}
\label{sec:con}
In this paper, we introduce a new setting of source-free unsupervised domain adaptation (SFUDA) for pulmonary nodule detection. Specifically, we aim to adapt the well-trained pulmonary nodule detector from the source domain to the unseen target domain without access to the private source data. We propose a novel Source-free Unsupervised cross-domain method for Pulmonary nodule detection, called Instance-level Contrastive Instruction fine-tuning framework (SUP-ICI). The source model is first adapted to the target domain via instance-level contrastive learning, where the RPN is employed to discriminate the nodules and other instances, and then trained in a teacher-student manner associated with a weighted entropy loss to further improve the detection performance on dealing with the noisy labels. Extensive experiments on the benchmark based on four public pulmonary nodule datasets demonstrate the effectiveness of our method. In the future, we plan to extend our method to the open set and open-world settings for medical applications, or investigate how to maintain the model performance on the source domain.




\bibliographystyle{IEEEtran}
\bibliography{main}

\end{document}